\documentclass [12pt]{article}
\usepackage{pslatex}
\usepackage{graphics}
\newtheorem{claims}{Assumption}

\begin{document}

\title{The Investment Acceleration Principle Revisited by Means of a Neural Network}
\author{Guido Fioretti \thanks{This research received financial support by \textit{Consiglio Nazionale delle Ricerche} and \textit{Deutsche Forschungsgemeinschaft}. E-mail: fioretti@cs.unibo.it.} \\ Universit\"{a}t Stuttgart \\ Institute for Theoretical Physics and Synergetics \\ Stuttgart, Germany \\ and \\ International Centre for Economic Research \\ Turin, Italy}
\maketitle

\begin{abstract}
The investment acceleration principle is a heuristic for modelling investment time series out of consumption time series. The model presented herein develops a disaggregated accelerator equation whose coefficients are the weights of a Kohonen neural net that represents firms' decision-making. According to this model, investments take place when managers recognise emerging technological patterns. Furthermore, a technique borrowed from the theory of self-organising systems is used in order to disentangle innovation-driven investments from plant-replication investments.
\end{abstract}


\section{Introduction}   \label{sec:introduction}

Investments in plants, machineries, scientific research and other forms of capital goods  are by far the most volatile, least predictable among economic time series. Yet this is precisely the magnitude that one would like to explicate in terms of a few determinants, since it is the engine of economic growth.

The \emph{investment acceleration principle} is a simple heuristic for modelling investment decision-making. Essentially, it asserts that investments depend on variations of demand.

In spite of its simplicity, the investment acceleration principle performs relatively well in empirical studies. A number of recent evaluations ascribe to accelerator equations a better predictive power than any competing model in such diverse situations as industrialised economies, developing economies and transition economies~\cite{bean81} \cite{abel-blanchard86} \cite{mckibbin-siegloff88} \cite{beng92} \cite{reiffers95} \cite{baxter96} \cite{bigsten-...99} \cite{mairesse-hall-mulkay99} \cite{lensink-sterken00}.  Not surprisingly, the investment acceleration principle is a major ingredient of many classical business cycle models such as Kaldor's~\cite{kaldor40}, Samuelson's~\cite{samuelson39}, Hicks's~\cite{hicks50}, Goodwin's~\cite{goodwin51} and Lucas's~\cite{lucas75}.

Business cycle models that rely on the investments acceleration principle generally make the following pair of assumptions:
\begin{claims}
Firms invest proportionally to variations of demand. \label{ass:variations}
\end{claims}
\begin{claims}
Firms only observe demand for their own products or, at most, for  products of their industrial sector. \label{ass:itsowndemand}
\end{claims}

Assumption~\ref{ass:variations} implies that firms do not make long-term strategic plans. It presumes a large degree of myopia in decision-making, as well as irrelevance of financial considerations.
Assumption~\ref{ass:itsowndemand} implies that the firms that produce
intermediate goods neither observe the final goods market, nor
any other indicator of the general state of the economy.
Thus, assumption~\ref{ass:itsowndemand} ascribes managers a sectoral myopia, in the sense that investment decisions are not influenced by consideration of trends in markets that are far down the productive chain.

Assumption~\ref{ass:variations} provides impulses that trigger the oscillations of economic
variables. In its turn, assumption~\ref{ass:itsowndemand} leads to formulate auto-regressive equations of  order $p\geq 2$
that are able to generate  oscillations of the macroeconomic variables. Parameter \(p\) is the number of
production stages, usually set at two.

I already mentioned that, although these assumptions are very crude, accelerator equations perform better than competing models based on financial considerations. However, even accelerator equations are far from yielding reliable predictions.

In order to understand the limits of the investment acceleration principle and why neural networks could be useful in order to improve on existing models, it is important to remark that the investment acceleration principle was originally formulated in order to describe the ascending phase of business cycles and the beginnings of crises~\cite{aftalion08-09}. The idea was that if firms react only to their own demand and if production of capital goods takes longer than production of final goods, then any increase of the demand for final goods  commands investments in capital goods, including temporary fluctuations. Thus, continuous increase of demand during the ascending phase of business cycles commands a continuous increase of utilised capital goods, but a small fluctuation of final demand translates into a stock of underutilised capital goods. This may be a big problem for the firms that made these investments, and the descending phase of the business cycle eventually begins.

Thus, correct timing of crises and duration of ascending phases can be improved by careful analysis of the commercial network and production times. However, the auto-regressive equations of order $p\geq 2$ that can be derived from assumptions~\ref{ass:variations} and~\ref{ass:itsowndemand} can generate whole business cycles, but the reason and the timing of recoveries is far less clear. In fact, the onset of economic recoveries
is due to investments in novel fields of activity
opened up by technological innovation. Thus, extending the investment acceleration principle to this part of economic dynamics requires microeconomic modelling of the cognitive process of recognition of novel investment opportunities~\cite{goodwin-punzo87}.

This paper attempts to do this by reproducing firms decision-making
by means of a self-organising map (SOM).
In the ensuing model, each neuron represents a firm with its 
ability to form categories out of information stemming both from demand
of given goods and from exogenous technological innovation. Neurons weights will correspond to the coefficients of a disaggregated accelerator equation, which can adapt its response to variations of demand according to  evolution of technological possibilities.

The paper is organised in two main sections, numbered~\ref{sec:neuralnet} and~\ref{sec:analdescr} respectively, followed by a concluding section~\ref{sec:conclusion}. Section~\ref{sec:neuralnet} illustrates the neural network and tests its behaviour with a simulation based on exogenous information regarding novel technologies. Section~\ref{sec:analdescr} attempts to reproduce its findings by means of an analytical description based on the theory of self-organising systems. Finally, section~\ref{sec:conclusion} concludes.

\section{The Neural Network}    \label{sec:neuralnet}

The minimal economic structure that we need to consider in order to apply the investment acceleration principle
involves households, firms that produce final goods
(henceforth labelled \textit{final goods sector}) and firms that
produce capital goods (henceforth labelled \textit{capital goods sector}). Production takes place in two stages and assumptions~\ref{ass:variations} and~\ref{ass:itsowndemand} of \S~\ref{sec:introduction} will be made.

Within this framework, 'investments' are purchases of capital
goods carried out by the firms that produce final goods. For 
simplicity, let us suppose a constant number of firms in 
both sectors.

Let us represent a firm's  decision-making
by means of:
\begin{itemize}
\item A first differences operator, yielding the variations of input variables (e.g. variations of demand).
\item A neuron, yielding a firm's purchase decisions based on 
recognition of the patterns entailed in input information. 
\item A summation operator, that integrates the neuron's output over time
in order to yield the current values of variables.
\end{itemize}

Input and output variables of our neural network are information vectors conveying orders for goods and services. Notably, information flows in the opposite direction of goods. 

Let vectors $\mathbf{c}$, $\mathbf{k}$, $\mathbf{l'}$ and $\mathbf{l''}$ denote households consumption, capital stock in the final goods sector, employment in the final goods sector and employment in the capital goods sector, respectively.  
Obviously, it must be
\(\mathbf{c} \geq \mathbf{0}\), \(\mathbf{k} \geq \mathbf{0}\), 
\(\mathbf{l'} \geq \mathbf{0}\), \(\mathbf{l''} \geq \mathbf{0}\).

Let vectors \(\Delta \mathbf{c}\), \(\Delta \mathbf{k}\),
\(\Delta \mathbf{l'}\) and \(\Delta \mathbf{l''}\) denote  variations of the above  vectors with respect to the previous time interval. These vectors carry information regarding demand for final goods, investments in capital goods and demand for labour in the final goods and capital goods sectors, respectively.

Furthermore, let a vector \(\mathbf{e}\) represent exogenous
information on innovations. Vector  \(\mathbf{e}\) does not represent
technological details that are developed by firms themselves
and that are kept strictly private unless acquired under
licensing agreements. Rather,  \(\mathbf{e}\) represents
all publicly available information about new technologies which
can induce managers to invest on a specific field, eventually
developing private information as a consequence of this decision.
It includes basic research made available by non-profit institutions,
rumours about competitors' strategies, as well as information that
was intended to be private but which is actually difficult to
appropriate and to trade, e.g. because of reverse engineering
\cite{arrow84}. Just like the investment accelerator principle assumes that firms react to \emph{variations} of demand, let us assume that firms react to variations $\Delta \mathbf{e}$ of the state of technology $\mathbf{e}$.

\begin{figure}
\center
\resizebox*{0.9\textwidth}{!}{\includegraphics{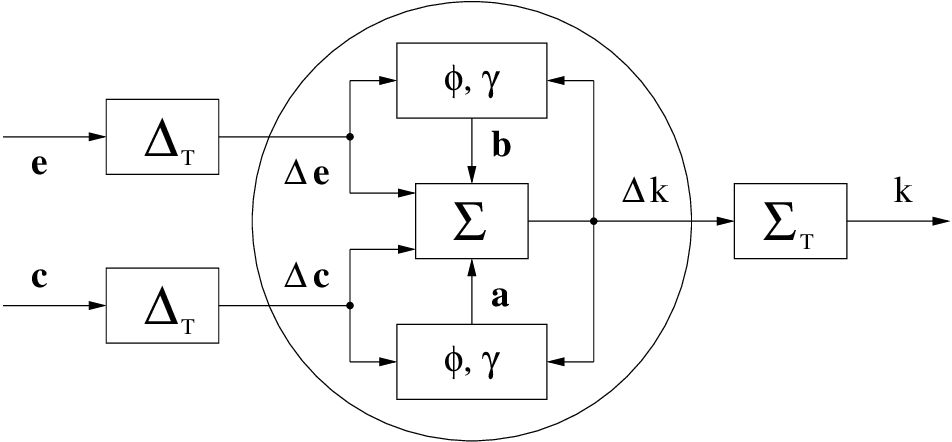}}
\caption{Decision-making by a firm in the final goods sector. The Kohonen neuron consists of a summator of inputs $\Delta c_{i}$ and $\Delta \mathbf{e}$ with weights $a_{i}$ and $b_{i}$, respectively. Both sets of weights are influenced by a feed-back $\phi$ and a feed-forward  $\gamma$, respectively. Left of the neuron, a first differences operator $\Delta$ performs the discrete-time derivative of the demand for final goods $\mathbf{c}$ and state-of-affairs in technology $\mathbf{e}$, respectively. Right of the neuron, a summator performs the discrete-time integration of investments $\Delta \mathbf{k}$, yielding capital stock $\mathbf{k}$.}  \label{fig:neuronDiffSumF}
\end{figure}

\begin{figure}
\center
\resizebox*{0.9\textwidth}{!}{\includegraphics{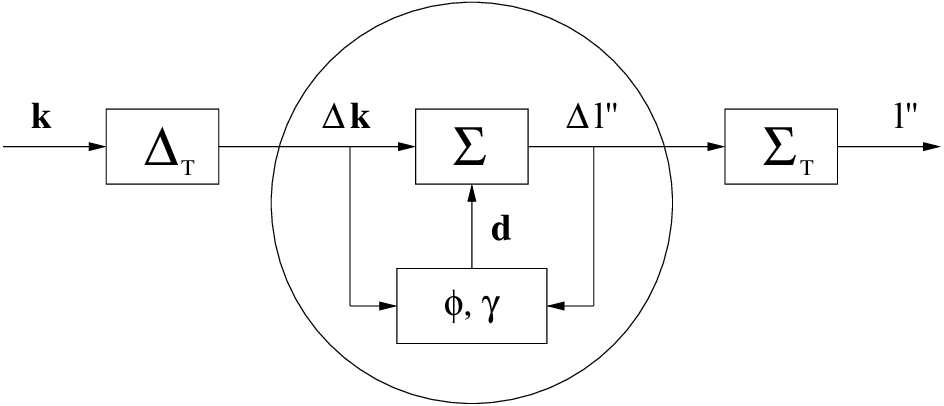}}
\caption{Decision-making by a firm in the capital goods sector. The Kohonen neuron consists of a summator of inputs $\Delta k_{i}$ with weights $d_{i}$. These weights are influenced by a feed-back $\phi$ and a feed-forward $\gamma$, respectively. Left of the neuron, a first differences operator $\Delta$ performs the discrete-time derivative of the demand for capital goods $\mathbf{k}$. Right of the neuron, a summator performs the discrete-time integration of demand for labour $\Delta \mathbf{l''}$, yielding employment $\mathbf{l''}$.} \label{fig:neuronDiffSumK}
\end{figure}

Let us suppose that the firms of the final goods sector react both to variation of demand and to variation of technologies, whereas the firms of the capital goods sector react to variations of demand only. Then, figures~\ref{fig:neuronDiffSumF} and~\ref{fig:neuronDiffSumK} illustrate decision-making by a firm of the final goods sector and by a firm of the capital goods sector, respectively. Demand for labour by the firms of the final goods sector does not appear in the figure because it is supposed to be proportional to demand for capital goods according to a law that will be specified subsequently. Demand for capital goods by the firms of the capital goods sector does not appear in the figure because this sector is supposed to include all production stages from mining to capital goods production.

The feed-backs and -forwards that characterise Kohonen neurons are of the utmost importance in order to model decentralised decision-making. On the contrary, a neural network where neurons weights would be set by an external operator would be  in order to represent a planned economy.

\begin{figure}
\center
\resizebox*{0.9\textwidth}{!}{\includegraphics{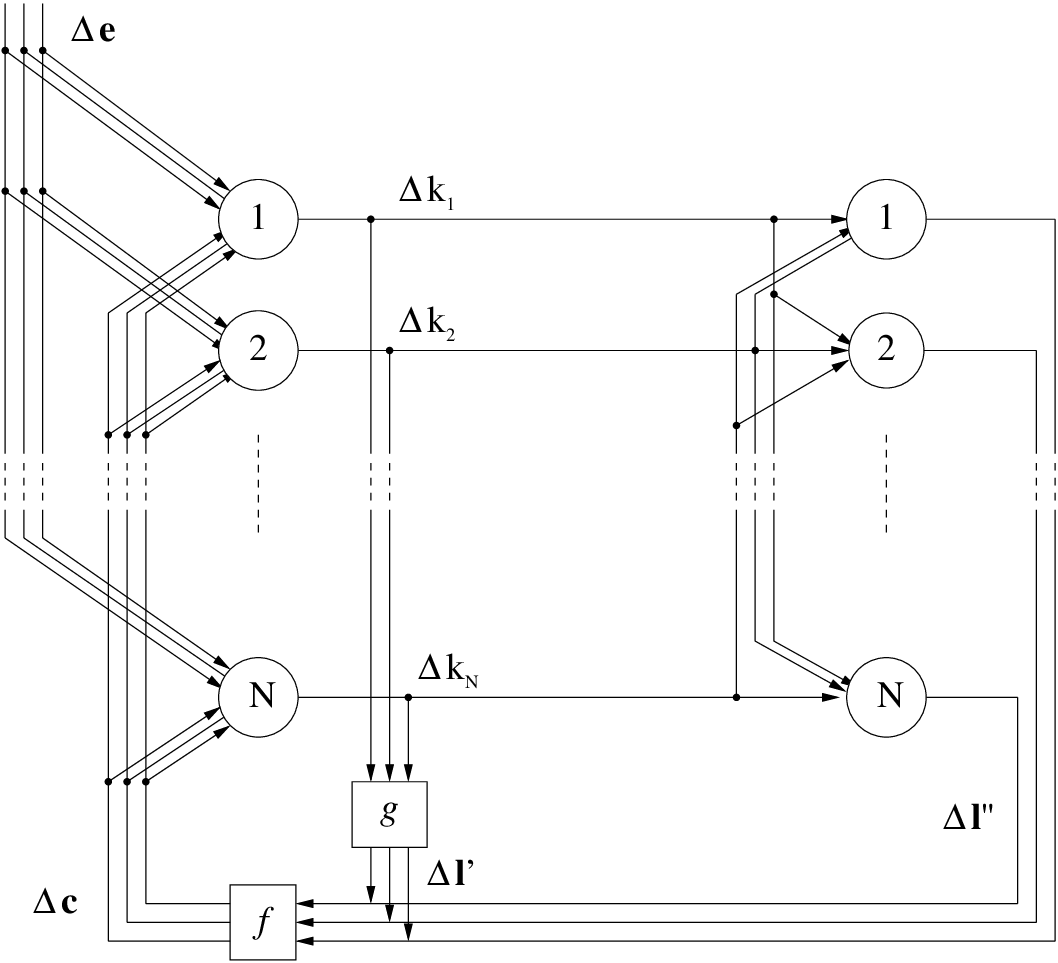}}
\caption{The structure of information flows within a productive system. Each neuron represents decision-making by one firm. The left layer represents firms in the final goods sector, the right layer represents firms in the capital goods sector, respectively. A double feed-back arises through the labour market.}    \label{fig:neuralnet}
\end{figure}

Let us arrange neurons in two layers, corresponding to firms
in the final goods sector and firms in the capital goods sector,
respectively. Figure~\ref{fig:neuralnet} illustrates the ensuing
neural network. Note that, since neurons are embedded in information circuits, sum and difference operators elide each other. Thus, in figure~\ref{fig:neuralnet} they have been ignored altogether.

Figure~\ref{fig:neuralnet} exhibits two 
boxes denoted  \(f\) and \(g\), respectively. Functions $f$ and $g$ represent
households behaviour and demand for labour by  the final goods
sector, respectively. These
aspects of decision-making do not need
to be modelled in detail in order to investigate investments acceleration. However, we must require that \(f(\Delta x)= \Delta f(x)\) and \(g(\Delta x)= \Delta g(x)\) in order to cancel sum and difference operators as above. Furthermore, note that since $f$ makes consumption depend on current income, we implicitly assumed that households do not save.

The network in figure~\ref{fig:neuralnet} has two nested feed-back loops, the inner one passing through the workers of the final goods
sector, the outer one through the workers of
the capital goods sector, respectively. Both feed-backs make demand depend on wages and, ultimately, on production in their respective sectors.

For simplicity, let us suppose that the final goods sector entails \(N\)
firms, that the capital goods sector entails \(N\) firms as well,  that there are \(N\) different final goods but only one capital good. In this way, all vectors will be \(N\)-dimensional. Note that although the number of
different goods is fixed with time, their qualitative features
may vary as a consequence of technological innovation.

Let an $N \times N$ matrix \(\mathbf{A}\) entail  the weights by which the neurons that represent decision-making by the final goods sector process information provided by demand. The \(i\)-th row of matrix \(\mathbf{A}\) entails the weights by which neuron \(i\)  multiplies \(\Delta \mathbf{c}\).

Similarly, let an $N \times N$ matrix \(\mathbf{B}\) entail the weights by which the neurons that represent decision-making by the final goods sector  process information provided by technological innovation. The \(i\)-th row of \(\mathbf{B}\) entails the weights by which neuron \(i\)  multiplies \(\Delta \mathbf{e}\).

Finally, let an $N \times N$ matrix \(\mathbf{D}\) entail the
weights by which the neurons that represent decision-making by the capital goods sector process information provided by demand for capital goods. The \(i\)-th row of  \(\mathbf{D}\) entails the weights by which neuron \(i\)  multiplies \(\Delta \mathbf{k}\).

Thus, information processing is described by:
\begin{eqnarray}
\Delta \mathbf{k}_{t} & = & \mathbf{A} \; \Delta \mathbf{c}_{t-1} \; + \:
\mathbf{B} \; \Delta \mathbf{e}_{t-1}                   \label{eq:AB}   \\
\Delta \mathbf{l}_{t}'' & = & \mathbf{D} \; \Delta \mathbf{k}_{t-1}   \label{eq:D}
\end{eqnarray}           
where 
$\Delta \mathbf{k}_{t} = \mathbf{k}_{t} - \mathbf{k}_{t-1}$,
$\Delta \mathbf{c}_{t-1} = \mathbf{c}_{t-1} - \mathbf{c}_{t-2}$,
$\Delta \mathbf{e}_{t-1} = \mathbf{e}_{t-1} - \mathbf{e}_{t-2}$,
$\Delta \mathbf{l}_{t}'' = \mathbf{l}_{t}'' - \mathbf{l}_{t-1}''$,
$\Delta \mathbf{k}_{t-1} = \mathbf{k}_{t-1} - \mathbf{k}_{t-2}$.

If we assume for simplicity that $f \equiv g \equiv \mathbf{I}$, we can write:
\begin{eqnarray}
\Delta \mathbf{c}_{t} & = & \Delta \mathbf{l}_{t}            \label{eq:f}   \\
\Delta \mathbf{l}_{t}' & = & \Delta \mathbf{k}_{t}           \label{eq:g}
\end{eqnarray}  
         
By combining equations~\ref{eq:AB} and~\ref{eq:D} with~\ref{eq:f} and~\ref{eq:g} and by remembering that \(\mathbf{l}= \mathbf{l'}+ \mathbf{l''}\),
we obtain a system of equations that takes information on technologies \(\Delta \mathbf{e}\) as
input and yields disaggregated investments \(\Delta \mathbf{k}\) as output. Eventually, by
summation of the components of \(\Delta \mathbf{k}\) one obtains aggregate investments.

Business cycle models make use of accelerator equations that, apart from being aggregate, are not different from a combination of equations~\ref{eq:AB}, \ref{eq:D}, \ref{eq:f} and~\ref{eq:g}. However, we are not interested in generating a business cycle but rather in providing a better modelling of the very beginning of its ascending phase. Thus,  we shall link the coefficients of these equations to the recognition of technological innovations.

The coefficients of equations~\ref{eq:AB} and~\ref{eq:D}  change according to the experiences and cognitive abilities of decision-makers. The equations that regulate the
evolution of coefficients matrices \(\mathbf{A}\), \(\mathbf{B}\), \(\mathbf{D}\) can take a number of forms \cite{kohonen89} \cite{kohonen95}. Typically, $\phi$ implements \emph{learning} by multiplying the inputs and the output of a neuron in order to stress those weights that yielded a high output with particular combinations of inputs. On the contrary, $\gamma$ implements \emph{forgetting} by decreasing the absolute value of weights with time, for instance with an exponential decay.

Let us choose the simplest ``Hebbian rule'' \cite{hebb49} out of the many equations proposed by
Kohonen~\cite{kohonen89}:
\begin{eqnarray}
\Delta \mathbf{A} & = & \mu \; \Delta \mathbf{k} \; \Delta \mathbf{c}^{T} - \nu  \, \mathbf{A}    
                                       \label{eq:Avariation} \\
\Delta \mathbf{B} & = & \mu \; \Delta \mathbf{k} \; \Delta \mathbf{e}^{T} - \nu \, \mathbf{B}    
                                       \label{eq:Bvariation} \\
\Delta \mathbf{D} & = & \mu \; \Delta \mathbf{l''} \, \Delta \mathbf{k}^{T} - \nu \, \mathbf{D}    
                                       \label{eq:Dvariation} 
\end{eqnarray}           

Equations~\ref{eq:Avariation}, \ref{eq:Bvariation}, and~\ref{eq:Dvariation} specify the variation of the coefficients of equations~\ref{eq:AB} and \ref{eq:D}, which in combination with equations~\ref{eq:f} and~\ref{eq:g} capture the essence of the investment acceleration principle. By combining accelerator equations~\ref{eq:AB}, \ref{eq:D}, \ref{eq:f}, \ref{eq:g} with coefficients variation equations~\ref{eq:Avariation}, \ref{eq:Bvariation}, \ref{eq:Dvariation} one obtains investment decisions $\Delta \mathbf{k}$ out of an exogenous sequence of information on innovations $\Delta \mathbf{e}$.

However, it seems sensible to constrain neurons as follows: 
\begin{enumerate}
\item The output of neurons is not allowed to become negative. This amounts to assume that capital is not scrapped and workers are not fired. \label{enu:one}
\item Credit exists, but loans cannot be indefinitely large. Since it is likely that capital stock serves as collateral, it is assumed that the output of a neuron cannot be larger than its cumulative output (but in order to allow start-ups to invest, this rule is not applied if cumulative output is zero). \label{enu:two}
\end{enumerate}

At this point, by combining the above  equations and rules and by feeding the model with a sequence of information vectors on technological innovations $\Delta \mathbf{e}$, one can run simulations that yield investments  $\Delta \mathbf{k}$. Since we want to model the very beginning of economic recoveries when patterns of novel technological possibilities emerge out of exhausted technologies, let us choose a sequence of vectors $\mathbf{e}$ such that a simple sinusoidal pattern emerges from noise.

\begin{figure}
\center
\resizebox*{0.9\textwidth}{!}{\rotatebox{-90}{\includegraphics{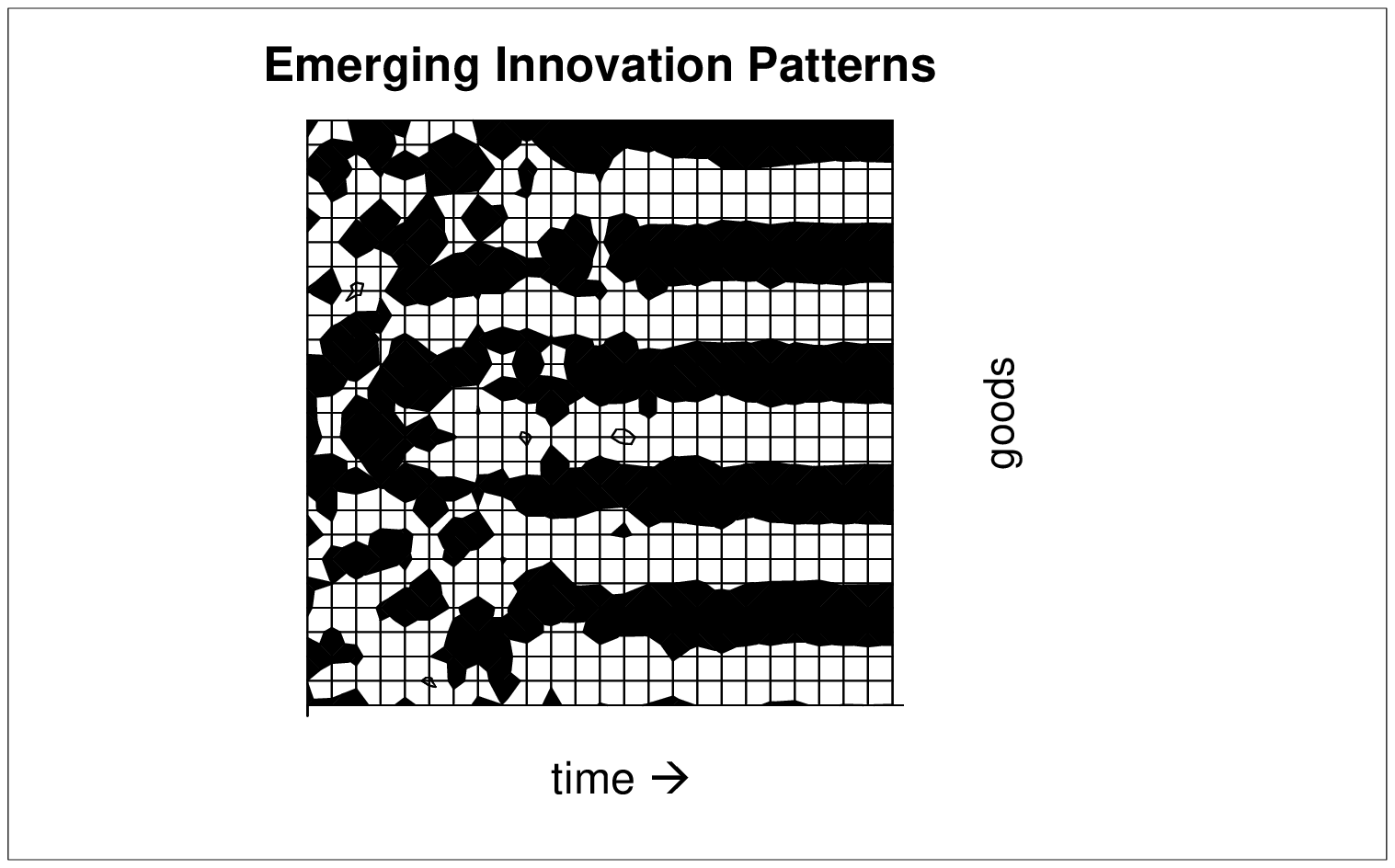}}}
\caption{A sequence of a hundred vectors $\mathbf{e}$ from $t=1$ to $t=100$, horizontal sections at $\mathbf{e}=\mathbf{0}$. Black areas denote regions where $\mathbf{e} > \mathbf{0}$, white areas denote regions where $\mathbf{e} \leq \mathbf{0}$. The sinusoid spans 100 goods with 5 periods of 20 goods each, its amplitude increasing from $A_{min}=0$ at $t=1$ to $A_{max}=2$ at $t=100$. Upon it, a normal distribution centred around zero adds noise with variance decreasing from $\sigma^{2}_{max}=1$ at $t=1$ to $\sigma^{2}_{min}=0$ at $t=100$. In order to simplify the image only one out of four goods and one out of four time periods have been shown, resulting in a $25 \times 25$ grid.}    \label{fig:e}
\end{figure}

A simple choice is that of having a sinusoid over goods whose amplitude increases over time from $A_{min}$ to $A_{max}$ at the same time that a normally distributed noise with zero mean decreases its variance down to zero. Figure~\ref{fig:e} illustrates a horizontal section of this combination drawn at $\mathbf{e}=\mathbf{0}$.

Vector $\Delta \mathbf{e}$ is obtained from $\mathbf{e}$ by taking a vector of zeros as initial condition. Since managers are likely to attach comparable importance to information on innovation and information on demand, the components of vectors $\Delta \mathbf{e}$ and $\Delta \mathbf{c}$ should be of similar amplitude. Thus, at each simulation step and before feeding $\Delta \mathbf{e}$ into the model, the following operations are carried out:
\begin{itemize}
\item The components of $\Delta \mathbf{e}$ that eventually exceed  the $[-A_{max}, A_{max}]$ interval, are cut.
\item The interval spanned by $\Delta \mathbf{e}$ is adjusted to the interval spanned by $\Delta \mathbf{c}$ by multiplying $\Delta \mathbf{e}$ by $[max(\Delta \mathbf{c}) - min(\Delta \mathbf{c})]/2A_{max}$.
\item The median of $\Delta \mathbf{e}$ is shifted to that of $\Delta \mathbf{c}$.
\end{itemize} 

\begin{figure}
\center
\resizebox*{0.9\textwidth}{!}{\rotatebox{-90}{\includegraphics{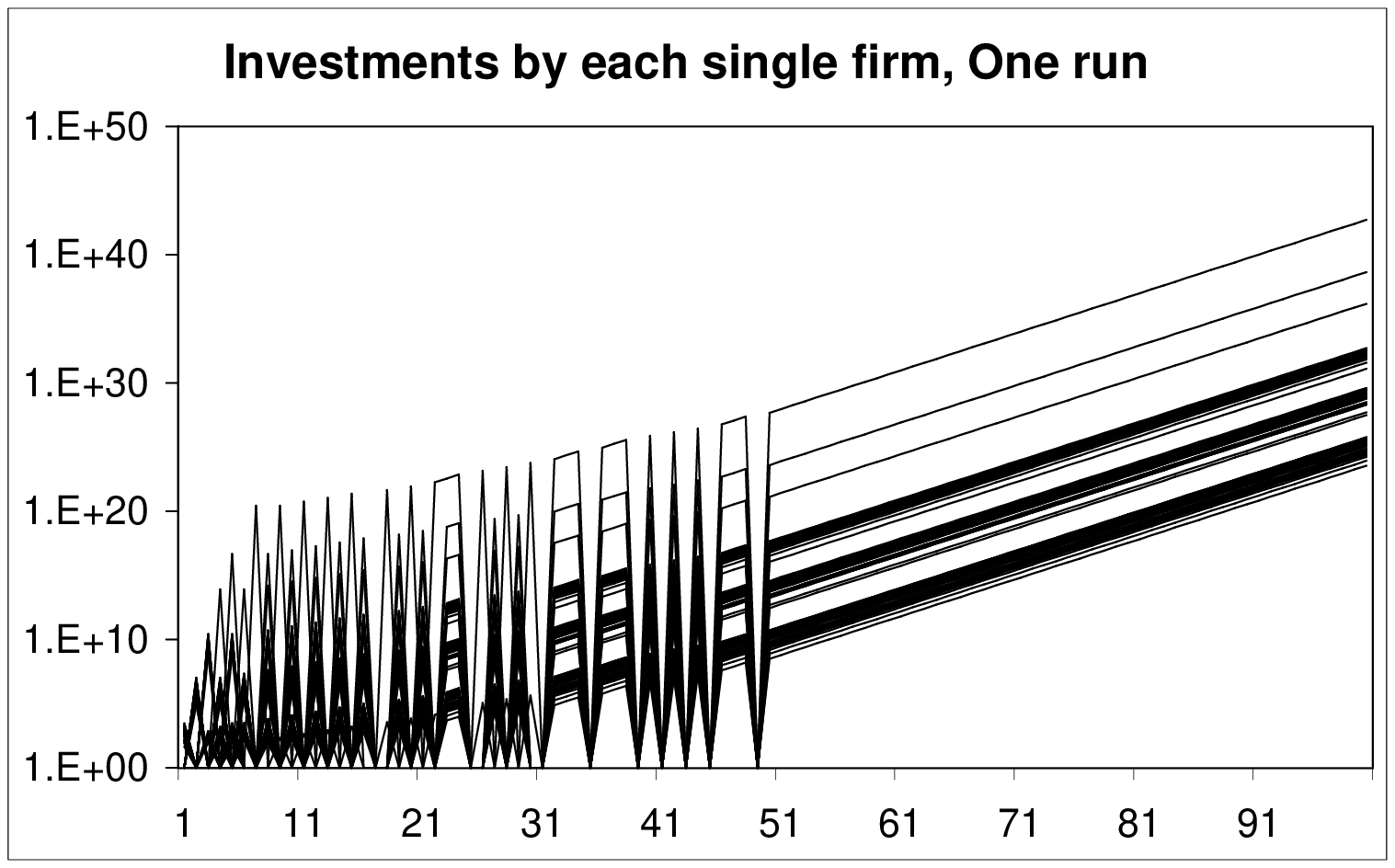}}}
\caption{Investments by each single firm. Each curve represents a different $\Delta k_{i}$ for $i=1,2,\ldots 100$ during  one single simulation.  In order to depict zero values of investments on the logarithmic scale, a one has been added to all values.}  \label{fig:disaggrInv}
\end{figure}

The resulting investments carried out by each single firm are illustrated in figure~\ref{fig:disaggrInv}. Each curve represents the output of one of the neurons of the left layer of figure~\ref{fig:neuralnet}. Initial conditions, keeping in mind that we are describing the onset of a recovery, are obviously $\mathbf{c}=\mathbf{0}$, $\mathbf{k}=\mathbf{0}$, $\mathbf{l'}=\mathbf{0}$, $\mathbf{l''}=\mathbf{0}$ and  $\Delta \mathbf{c}=\mathbf{0}$, $\Delta \mathbf{k}=\mathbf{0}$, $\Delta \mathbf{l'}=\mathbf{0}$, $\Delta \mathbf{l''}=\mathbf{0}$. Learning and forgetting parameters have been set at $\mu=0.1$ and $\nu=0.1$, respectively.

Since investments grow exponentially, logarithmic pictures have been used. In order to depict zero values, a one has been added to all components of $\Delta \mathbf{k}$.

The most notable feature of the graph illustrated in figure~\ref{fig:disaggrInv} is that at about a half of running time, precisely when according to figure~\ref{fig:e} a technological pattern is emerging, firms stop to invest erratically and step onto an exponential growth path.

\begin{figure}
\center
\resizebox*{0.9\textwidth}{!}{\rotatebox{-90}{\includegraphics{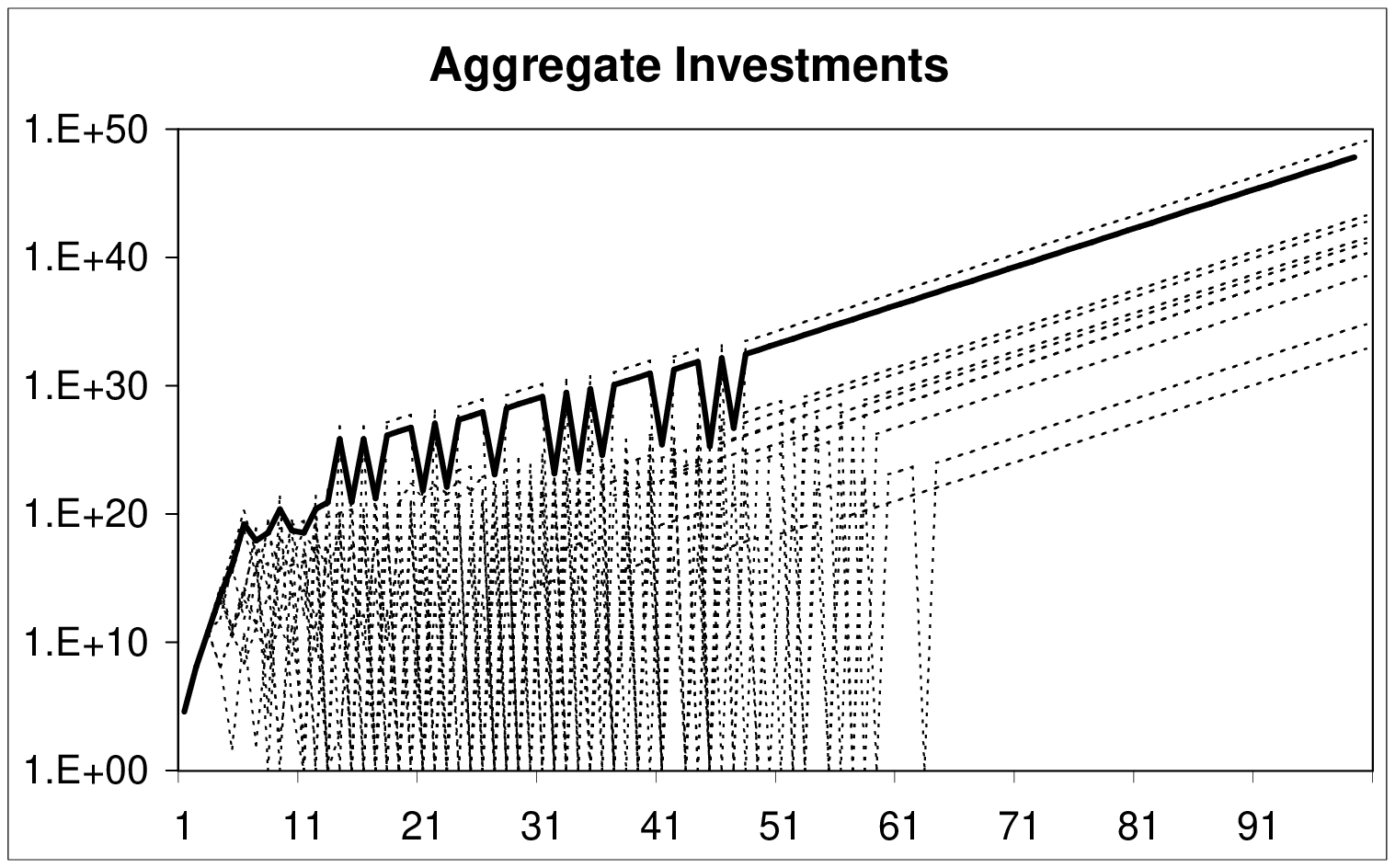}}}
\caption{Aggregate investments produced by ten different simulations (dashed lines) and their average (thick line).  In order to depict zero values of investments on the logarithmic scale, a one has been added to all curves.}  \label{fig:aggrInv}
\end{figure}

Other simulations produced qualitatively similar results. Figure~\ref{fig:aggrInv} illustrates aggregate investments produced by ten different simulations, as well as their average. Note that according to \emph{all} simulations  a stable exponential starts at some point between the $50^{th}$ and the $60^{th}$ time step. 

\begin{figure}
\center
\resizebox*{0.9\textwidth}{!}{\rotatebox{-90}{\includegraphics{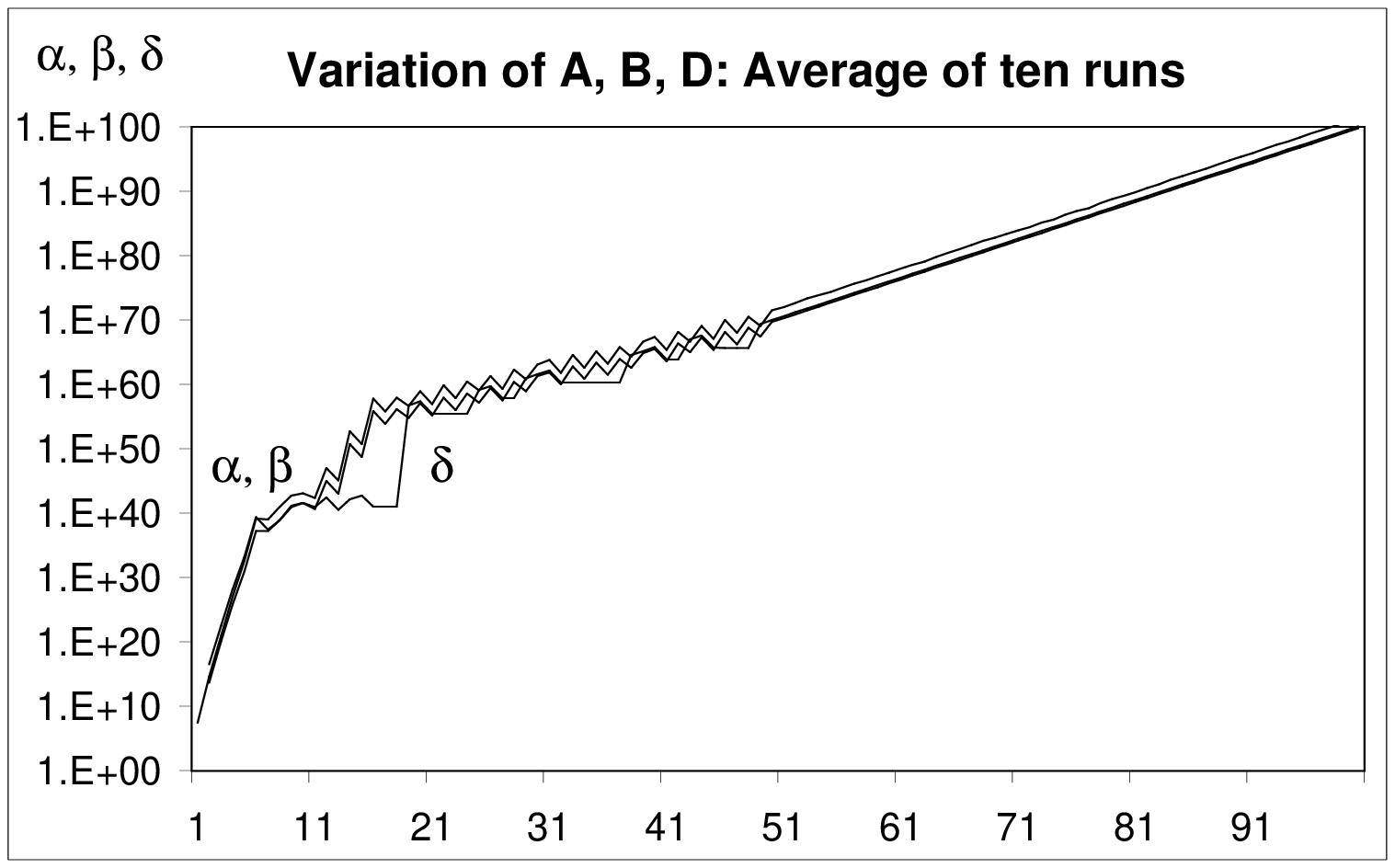}}}
\caption{Variation of matrices $\mathbf{A}$, $\mathbf{B}$, $\mathbf{D}$ illustrated by indicators $\alpha$, $\beta$, $\delta$, respectively. Indicators $\alpha$, $\beta$, $\delta$ are defined as the sum of the absolute variations of all elements of $\mathbf{A}$, $\mathbf{B}$, $\mathbf{D}$, respectively. The graph illustrates the average of ten simulations.}  \label{fig:deltaABD}
\end{figure}

Clearly, this happens because matrices $\mathbf{A}$, $\mathbf{B}$ and $\mathbf{D}$ change their weights. Figure~\ref{fig:deltaABD} illustrates three indicators of the variation of  $\mathbf{A}$, $\mathbf{B}$ and $\mathbf{D}$, respectively. It is clear that the recognition of innovations terminates a period of turbulence and sets the variation of $\mathbf{A}$, $\mathbf{B}$ and $\mathbf{D}$ on an exponential path that reflects the exponential growth of investments.

The above curves have been obtained out of a model that combines the variation of economic magnitudes such as demand, investments and information on technologies with the variation of cognitive parameters represented by $\mathbf{A}$, $\mathbf{B}$ and $\mathbf{D}$. On the contrary, the ensuing section makes an attempt to disentangle these two effects.

\section{Analytical Description}              \label{sec:analdescr}

Even if we assumed that \(f \equiv g \equiv \mathbf{I}\), any combination of
the remaining equations~\ref{eq:AB}, \ref{eq:D}, \ref{eq:f}, \ref{eq:g} 
is non-linear. This happens because  variables \(\Delta \mathbf{c}\),
\(\Delta \mathbf{k}\), \(\Delta \mathbf{l}\), \(\Delta \mathbf{e}\)
multiply variable coefficients \(\mathbf{A}\), 
\(\mathbf{B}\), \(\mathbf{D}\).

However, analytical treatment is possible on the ground that, since mental
categories must vary at a much slower pace than the information that
they classify, the corresponding weight matrices \(\mathbf{A}\),
\(\mathbf{B}\), \(\mathbf{D}\) must vary at a much slower pace than information vectors
\(\Delta \mathbf{c}\), \(\Delta \mathbf{k}\), \(\Delta \mathbf{l}\) and \(\Delta \mathbf{e}\). Furthermore, since classifying information in a mental category translates into a vector falling into a basin of attraction, fast information vectors are likely to be \emph{stable} as well.

Since $\Delta \mathbf{e}$ is an exogenous variable and since $\Delta \mathbf{l}$ can be expressed in terms of $\Delta \mathbf{c}$ and $\Delta \mathbf{k}$, let us focus on fast variables $\Delta \mathbf{c}$, $\Delta \mathbf{k}$ and slow variables $\mathbf{A}$ and $\mathbf{D}$ only. For the above reasons, fast variables $\Delta \mathbf{c}$ and $\Delta \mathbf{k}$ are stable within the basins of attraction specified by the slow variables $\mathbf{A}$ and $\mathbf{D}$. Thus, $\mathbf{A}$ and $\mathbf{D}$ are also called  \emph{order parameters}.

In their turn, the slow variables $\mathbf{A}$ and $\mathbf{D}$ evolve under the influence of the fast variables $\Delta\mathbf{c}$ and $\Delta\mathbf{k}$, respectively. In fact, innovation produces novel patterns that must be
classified by means of novel categories, so the basins of attraction change with time. Consequently, the slow variables $\mathbf{A}$ and $\mathbf{D}$ are inherently unstable.

This dichotomy between fast, stable variables and slow, unstable variables is typical of self-organising systems \cite{nicolis-prigogine77} \cite{haken87} \cite{nicolis-prigogine89} \cite{haken91}. Generally speaking, self-organisation can be studied analytically by separating:
\begin{enumerate}
\item A regime where fast variables
\(\Delta \mathbf{c}\) and \(\Delta \mathbf{k}\) are infinitely fast so they reach instantaneously their equilibrium values. In this situation, it is easy to study the evolution of slow variables \(\mathbf{A}\) and \(\mathbf{D}\).                                 \label{reg:first}
\item A regime where slow variables \(\mathbf{A}\) and \(\mathbf{D}\) are infinitely
slow, i.e. they are fixed coefficients. In this situation, it is easy to study the evolution of fast variables
\(\Delta \mathbf{c}\) and \(\Delta \mathbf{k}\).                    \label{reg:second}
\end{enumerate}

In our context, regime~\ref{reg:first} is approximated by the very
beginning of economic recoveries. In fact, in this phase investments are small, easy to carry out and, possibly, fast enough to be close to their equilibrium level. At the same time, a lot of technological novelties force managers to change their mental categories, represented by slow and unstable order parameters.

On the contrary,
regime~\ref{reg:second} is approximated by the growing phase
of the business cycle, when technologies vary less rapidly, the accelerator
coefficients are stable and large-scale investments take place
by replication of existing plants. In this phase, order parameters can be taken to be constant while investments grow exponentially according to traditional, fixed-coefficient accelerator equations.

When an economy is in regime~\ref{reg:first}, neurons processing
time can be neglected. Thus, upon examination of 
figure~\ref{fig:neuralnet} and by assuming \(f\equiv g \equiv
\mathbf{I}\) we can write \(\Delta \mathbf{l'} = \Delta \mathbf{k}\),
\(\Delta \mathbf{l''} = \mathbf{D} \, \Delta \mathbf{k}\),
\(\Delta \mathbf{c} = (\mathbf{I} + \mathbf{D}) \, \Delta \mathbf{k}\),
\(\Delta \mathbf{k} = \mathbf{A} \, \Delta \mathbf{c} +
\mathbf{B} \, \Delta \mathbf{e}\). Combination of these expressions yields:
\begin{equation}
\Delta \mathbf{k} \; = \; \frac{\mathbf{B}}{\mathbf{I} - \mathbf{A} -
\mathbf{A}\mathbf{D}} \: \Delta \mathbf{e}      \label{eq:firstreg}
\end{equation}

Thus, we can envision an alternative algorithm in order to derive a sequence  $\{ \Delta \mathbf{k} \}$ from a sequence $ \{ \Delta \mathbf{e} \}$. First, for each $\Delta \mathbf{e}$ equation~\ref{eq:firstreg} yields a vector $\Delta \mathbf{k}$ to which rules~\ref{enu:one} and~\ref{enu:two} of \S~\ref{sec:neuralnet} are applied in order to obtain economically feasible values. Subsequently, coefficient matrices $\mathbf{A}$ and $\mathbf{D}$ are updated by means of equations~\ref{eq:Avariation} and \ref{eq:Dvariation}.

\begin{figure}
\center
\resizebox*{0.9\textwidth}{!}{\rotatebox{-90}{\includegraphics{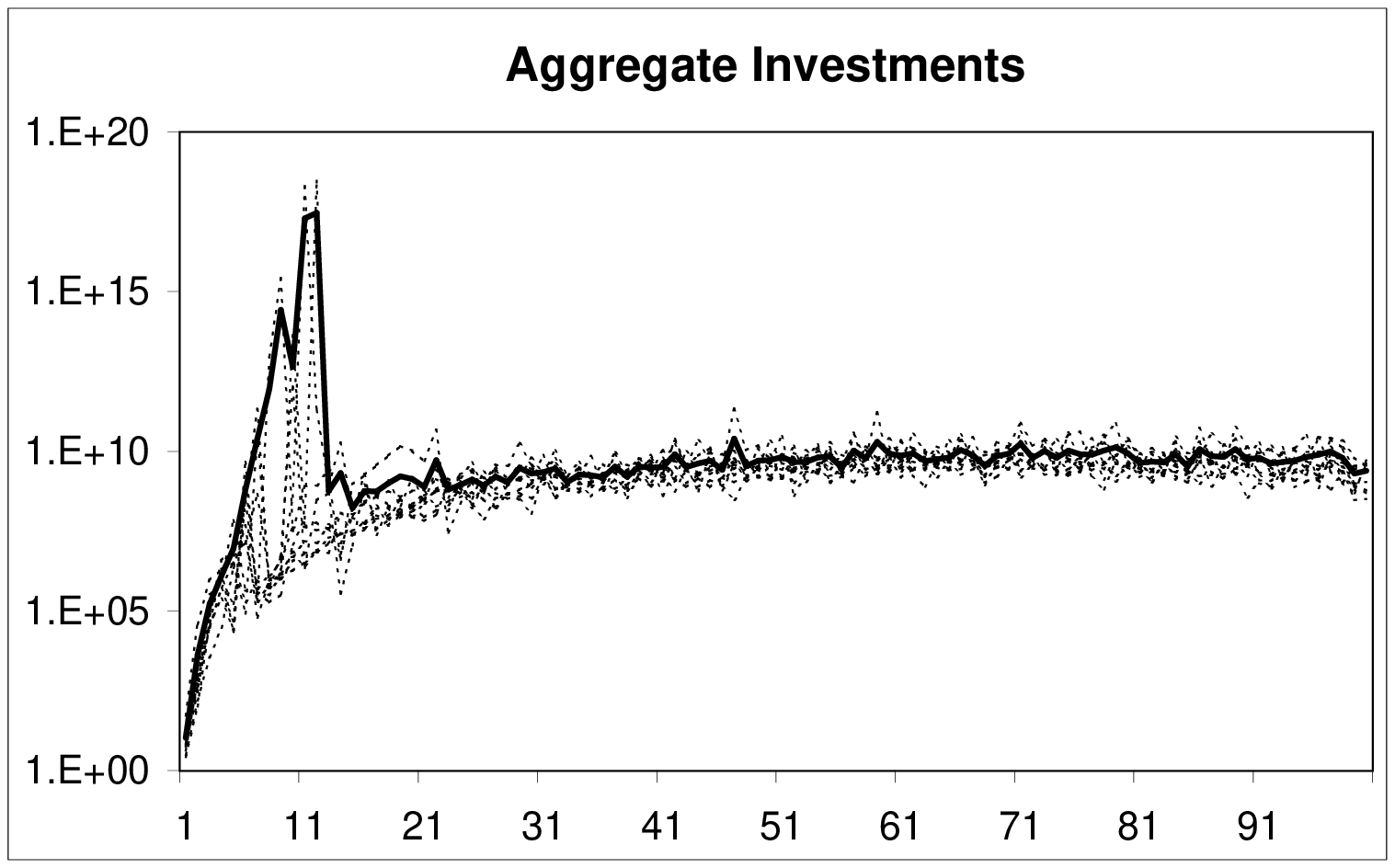}}}  \label{fig:instantAggrInv}
\caption{Aggregate investments  calculated by assuming that information vectors $\Delta\mathbf{c}$ and $\Delta\mathbf{k}$ vary with infinite speed. The figure illustrates ten single runs (dashed lines) as well as their average (thick solid line).}
\end{figure}

Figure~\ref{fig:instantAggrInv} illustrates aggregate investments calculated in this way over ten simulation runs, as well as their average. It is evident that, after a brief transitory, investments take on a constant value.

Figure~\ref{fig:instantAggrInv} makes clear that, contrary to the model expounded in Section~\ref{sec:neuralnet}, investments approach a constant, low level. This is hardly surprising, since assuming infinite speed of information takes away a basic assumption of the model, namely that production time lags and the structure of information flows between industrial sectors operating at different production stages may cause excessive productive capacity. However, the purpose of this infinite-speed model is only that of obtaining values of $\mathbf{A}$ and $\mathbf{D}$ that can be used in a standard, fixed-coefficients accelerator equation.

\begin{figure}
\center
\resizebox*{0.9\textwidth}{!}{\rotatebox{-90}{\includegraphics{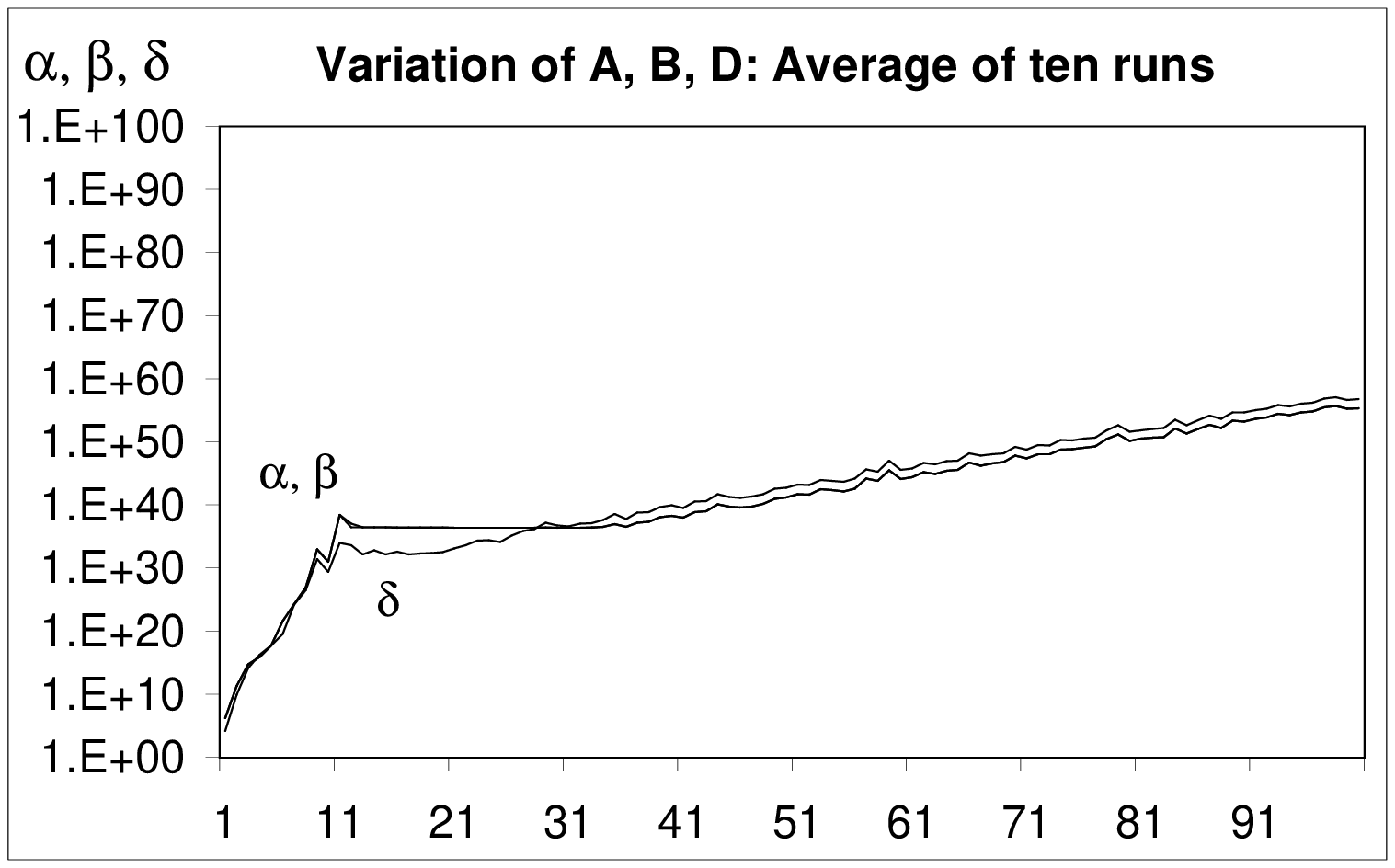}}}
\caption{Variation of matrices $\mathbf{A}$, $\mathbf{B}$, $\mathbf{D}$ illustrated by indicators $\alpha$, $\beta$, $\delta$, respectively. Indicators $\alpha$, $\beta$, $\delta$ are defined as the sum of the absolute variations of all elements of $\mathbf{A}$, $\mathbf{B}$, $\mathbf{D}$, respectively. The graph illustrate the averages of ten simulations.}  \label{fig:instantDeltaABD}
\end{figure}

Figure~\ref{fig:instantDeltaABD} illustrates the variation of $\mathbf{A}$, $\mathbf{B}$ and  $\mathbf{D}$ by means of the same indicators that had been used in \S~\ref{sec:neuralnet}. According to this figure, since the $30^{th}$ time step these matrices begin to move on a smooth exponential path.

Figure~\ref{fig:instantDeltaABD} makes clear that matrices $\mathbf{A}$, $\mathbf{D}$, $\mathbf{B}$, on average, approach a stable path as soon as the technological pattern begins to appear at the $30^{th}$ time step. Matrices $\mathbf{A}$, $\mathbf{D}$, $\mathbf{B}$ are still growing because equations~\ref{eq:Avariation}, \ref{eq:Bvariation} and~\ref{eq:Dvariation} make their variation proportional to the level of investments, but a stable variation like this could be easily inserted in an investments acceleration model that does not employ a neural network. The only difference with a fixed-coefficients model would be a more steeply exponential growth path.

Thus, let us use the values attained by $\mathbf{A}$ and $\mathbf{D}$ at the $30^{th}$ step  to run a traditional, fixed-coefficients accelerator equation that receives no exogenous input $\Delta \mathbf{e}$. In this equation, the initial level of investments will be the one attained when $\mathbf{A}$ and $\mathbf{D}$  were frozen. Note that, by doing this, we are entering regime~\ref{reg:second}.

By combining equation~\ref{eq:AB} deprived of the \(\mathbf{D} 
\Delta \mathbf{e}_{t-1}\) term with
 equations~\ref{eq:D}, \ref{eq:f}, \ref{eq:g}, 
one obtains:
\begin{equation}
\Delta \mathbf{k}_{t} \;\: = \;\: \mathbf{\tilde{A}} \, f(g(\Delta \mathbf{k}_{t-1}) \; +
\;\, \mathbf{\tilde{D}} \, \Delta \mathbf{k}_{t-2})                 \label{eq:secondreg}
\end{equation}
where $\mathbf{\tilde{A}}$ and $\mathbf{\tilde{D}}$ are the once variable~$\mathbf{A}$ and~$\mathbf{D}$ frozen at the $30^{th}$ time step.

Albeit disaggregated, equation~\ref{eq:secondreg} is a traditional, fixed-coefficients accelerator equation. Just as in the case of regime~\ref{reg:first}, each time step we have to check whether the outputs of the neurons are yielding feasible values by applying rules~\ref{enu:one} and~\ref{enu:two} of \S~\ref{sec:neuralnet}. 

\begin{figure}
\center
\resizebox*{0.9\textwidth}{!}{\includegraphics{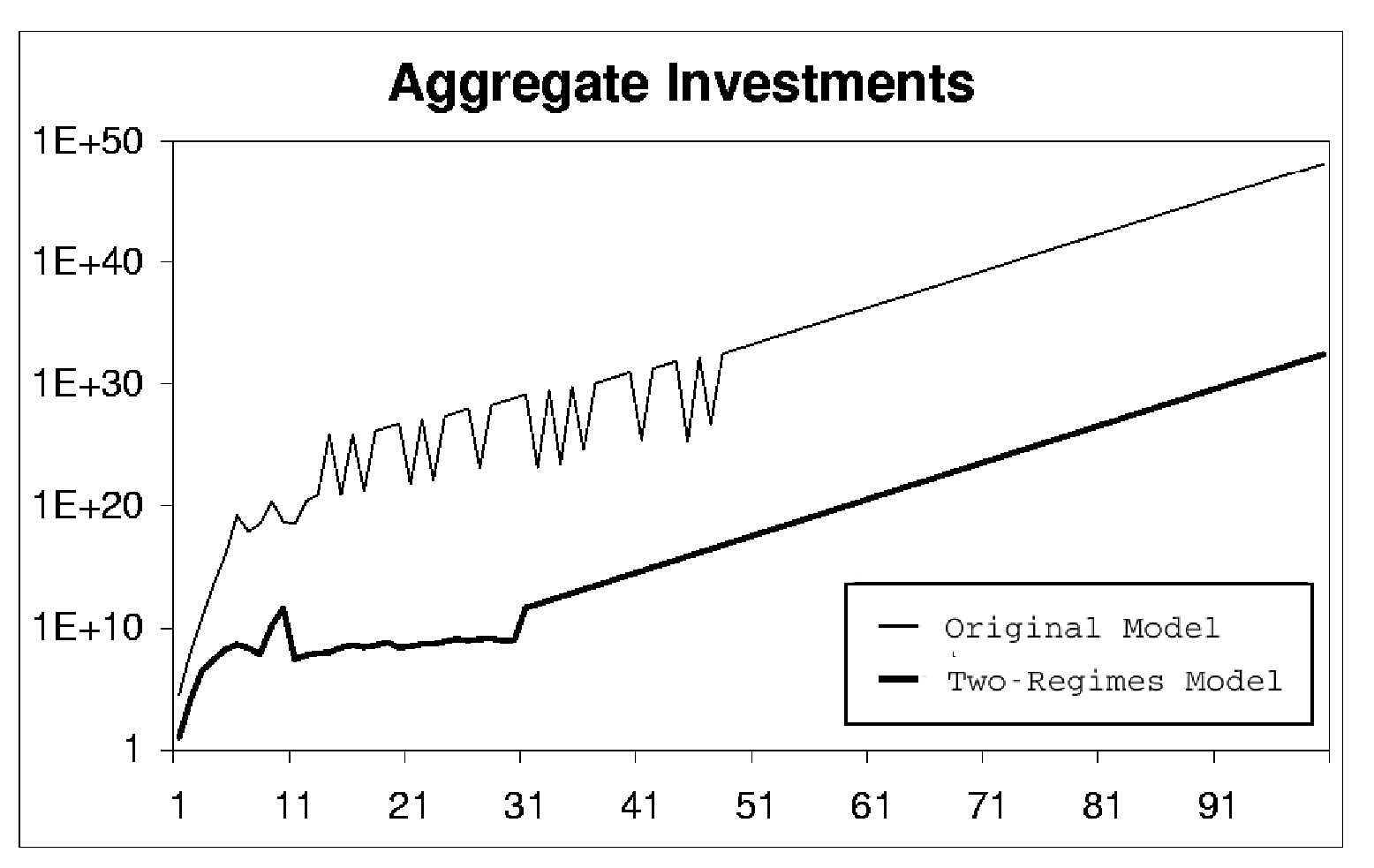}}
\caption{Aggregate investments according to the original model  (thin line) and aggregate investments according to the two-regimes model (thick line). Both curves depict the average of ten simulation runs.}  \label{fig:invComparison}
\end{figure}

Figure~\ref{fig:invComparison} compares the aggregate investments generated by the original model of \S~\ref{sec:neuralnet} with those of the two-regimes model expounded in this section. The graph refers to the averages of ten simulations.

Figure~(\ref{fig:invComparison}) allows us to compare a model where the two processes of understanding technological novelties and undertaking investments take place at the same time, with a model that separates these two concepts. It makes clear that traditional, fixed-coefficients accelerator equations are only able to account for investments that are carried out according to a given technological paradigm, once this emerged. In an earlier phase, where it is still unclear which will be the winning technologies, a neural network improves performance.

\section{Concluding Remarks} \label{sec:conclusion}

Although feed-forward neural networks find a lot of applications in handling financial data, the usage of a Kohonen neural networks in order to reproduce decision-making by economic agents is entirely new to economics. In this sense, this paper aimed to open a new field of research  by establishing a theoretical bridge between an economic model and a tool of artificial intelligence. 

This is theoretically interesting  because it allowed at least a partial modelling of a very important but very vague concept like that of ``corporate culture''. However, it has practical implications as well.

Firstly, Kohonen neural networks can provide a better understanding of the empirical dynamics of investments. A first attempt in this direction gave encouraging results \cite{fioretti-sussmuth01}.

Secondly, and possibly most importantly, Kohonen neural nets might be used on POS data in order to guide investments into novel fields. This is an entirely unexplored field of research, since until now POS data have been only used in order to forecast seasonal variations of demand at selling points or, at most, demand at a new location.

\bibliographystyle{plain}
\bibliography{accelInform}
\end{document}